\documentclass[11pt]{article}

\usepackage[final]{acl}

\usepackage{times}
\usepackage{latexsym}

\makeatletter
\def\@fnsymbol#1{\ensuremath{\ifcase#1\or \dagger\or *\or \ddagger\or
   \mathsection\or \mathparagraph\or \|\or \dagger\dagger
   \or \ddagger\ddagger \else\@ctrerr\fi}}
\makeatother

\usepackage[T1]{fontenc}

\usepackage[utf8]{inputenc}

\usepackage{microtype}

\usepackage{inconsolata}

\usepackage{graphicx}
\usepackage[misc]{ifsym}

\usepackage{amsmath}
\usepackage{amssymb}
\usepackage{algorithm}
\usepackage{algpseudocode}

\usepackage{xspace}
\usepackage{booktabs}
\usepackage{multirow}
\usepackage{array}
\usepackage{subcaption}
\usepackage{xcolor}
\usepackage{colortbl}

\usepackage[breakable]{tcolorbox} 
\usepackage{inconsolata}        

\newcommand{\approach}{{GlimpRouter}\xspace}

\title{GlimpRouter: Efficient Collaborative Inference by Glimpsing \\
One Token of Thoughts}

\author{
  Wenhao Zeng\textsuperscript{\dag}\quad
  Xuteng Zhang\textsuperscript{\dag} \quad
  Yuling Shi \quad
  Chao Hu \quad \\
  \textbf{Yuting Chen} \quad
  \textbf{Beijun Shen} \quad
  \textbf{Xiaodong Gu}\textsuperscript{{\textrm{\Letter}}} \\
  Shanghai Jiao Tong University \quad \\
  \texttt{\{zengwh\_cs, scottzhang, xiaodong.gu\}@sjtu.edu.cn}
}

\begin{document}
\maketitle
{\renewcommand{\thefootnote}{}
\footnotetext{\textsuperscript{\dag}Equal contribution. {\textrm{\Letter}}\,Corresponding author.}}
\begin{abstract}
Large Reasoning Models (LRMs) achieve remarkable performance by explicitly generating multi-step chains of thought, but this capability incurs substantial inference latency and computational cost. Collaborative inference offers a promising solution by selectively allocating work between lightweight and large models, yet a fundamental challenge remains: determining when a reasoning step requires the capacity of a large model or the efficiency of a small model. Existing routing strategies either rely on local token probabilities or post-hoc verification, introducing significant inference overhead.
In this work, we propose a novel perspective on step-wise collaboration: the difficulty of a reasoning step can be inferred from its very first token. Inspired by the ``Aha Moment'' phenomenon in LRMs, we show that the entropy of the initial token serves as a strong predictor of step difficulty. 
Building on this insight, we introduce \approach, a training-free step-wise collaboration framework. \approach employs a lightweight model to generate only the first token of each reasoning step and routes the step to a larger model only when the initial token entropy exceeds a threshold. Experiments on multiple benchmarks demonstrate that our approach significantly reduces inference latency while preserving accuracy. For instance, \approach attains a substantial 10.7\% improvement in accuracy while reducing inference latency by 25.9\% compared to a standalone large model on AIME25. These results suggest a simple yet effective mechanism for reasoning: allocating computation based on a glimpse of thought rather than full-step evaluation
\footnote{Code and dataset are available at \url{https://github.com/Zengwh02/GlimpRouter}.}.
\end{abstract}



\section{Introduction}
\label{sec:introduction}

Large reasoning models, such as DeepSeek-R1~\cite{guo2025deepseek} and OpenAI o1/o3~\cite{jaech2024openai, openai2025o3}, have demonstrated remarkable performance across a wide range of complex reasoning tasks by explicitly generating structured reasoning steps~\cite{yang2025elaboration,wang2025vrag,wang2025vidorag,chen2025swe,li2025swe,hu2025flowmaltrans,liu2025attention,shi2024between}. However, this capability comes at a high cost: extended reasoning chains inevitably incur high inference latency and substantial computational overhead, severely limiting the practicality of LRMs in latency-sensitive and resource-constrained settings.

To alleviate this bottleneck, collaborative inference has emerged as a promising paradigm~\cite{she2025hawkeye, xiao2025proxythinker, qu2025survey}. The key idea is that multiple models with heterogeneous capabilities and costs are orchestrated to jointly solve a task: lightweight models handle routine or easy tasks, while more powerful but expensive models are invoked selectively for difficult ones~\cite{chen2025survey, wang2025mixllm}. 
Existing collaboration strategies generally fall into two categories. Token-level methods, such as Speculative Decoding~\cite{leviathan2023fast, fu2025r2r}, accelerate generation by allowing a small model to propose candidate tokens that are then verified (and possibly accepted) by a larger model, reducing the number of expensive forward passes. 
In contrast, step-level methods attempt to route entire reasoning steps between models. They typically depend on \textit{post-hoc} verification, such as LLM-as-a-judge~\cite{shi2025speccot,pan2025specreason}, or on averaged uncertainty metrics (e.g., mean perplexity or entropy)~\cite{cui2025stepwise, zhang2025entropy}.

Despite their promise, determining the optimal allocation of tasks between large and small models remains a fundamental challenge.
Token-level methods rely primarily on local token probabilities and frequently switch models at a fine granularity, while step-level strategies require evaluating entire reasoning steps before making routing decisions. 
Both approaches introduce non-trivial computational overhead, which can partially or even fully offset their intended efficiency gains. 

In this work, we advance a new perspective: \textit{Can we judge the difficulty of a reasoning step at the very beginning?} Inspired by the ``Aha Moment'' phenomenon observed in LRMs~\cite{guo2025deepseek}, previous works suggest that the onset of a reasoning step--often marked by discourse cues such as ``Wait'', ``But'', or ``So''--represents a cognitive pivot that shapes the entire subsequent reasoning trajectory~\cite{wang2025beyond,yang2025speculative,zeng2025pruning}. We hypothesize that the information associated with this \textit{initial token} provides a more discriminative signal of step-level difficulty than averaged statistics over the full step.
To validate this hypothesis, we analyze the uncertainty distribution of the reasoning steps across various tasks (Section~\ref{sec:motivation}). The experimental results show that the entropy of the initial tokens exhibits an exceptionally high variance, indicating it is a powerful discriminator: steps with low initial entropy typically correspond to routine derivations that a small model can readily process, whereas steps with high initial entropy indicate critical cognitive bifurcations that require the collaboration of larger, more powerful models.

Building on this insight, we propose \textbf{\approach}, a novel training-free framework for efficient collaborative inference. At each reasoning step, a lightweight model is first used to generate only the initial token of the step. If the entropy of this token falls below a predefined threshold, the lightweight model proceeds to generate the entire step; otherwise, the generated context is seamlessly handed over to a larger and more capable model. In effect, this mechanism essentially operationalizes the principle of ``\textit{A Glimpse of Thought}'': a minimal signal from the onset of a reasoning step governs the computational budget allocated to its full generation.

We empirically evaluate our approach on a diverse set of reasoning benchmarks, including mathematical reasoning (AIME), challenging general reasoning (GPQA~\cite{rein2024gpqa}), and code generation (LiveCodeBench~\cite{jain2024livecodebench}). 
Across all benchmarks, \approach achieves a superior trade-off between efficiency and performance. 
For instance, on the AIME25 benchmark, our method attains a substantial \textbf{10.7\%} improvement in accuracy while simultaneously reducing inference latency by \textbf{25.9\%} compared to a standalone large model. Moreover, we demonstrate that our step-level routing strategy is orthogonal to token-level speculative decoding, enabling complementary and compound speedups when integrated.

Our contributions are summarized as follows:
\begin{itemize}
    \setlength{\itemsep}{0pt}
    \item We perform an analysis of the uncertainty distribution across reasoning steps and identify that the entropy of the initial token serves as an indicator of reasoning difficulty.
    \item We propose \textbf{\approach}, a simple yet effective mechanism that enables training-free, step-aware collaboration between models.
    \item Extensive experiments verify that our method significantly reduces latency while maintaining or even enhancing the reasoning efficacy of large models, offering a practical solution for deploying efficient LRMs.
\end{itemize}

\begin{figure*}[t]
    \centering
    \includegraphics[width=\textwidth]{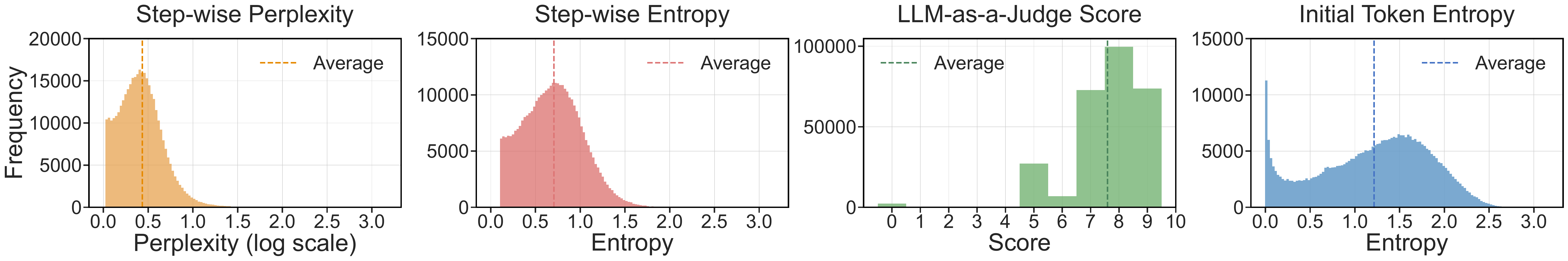}
    \caption{
    Comparison of the distributions of different uncertainty metrics. Unlike other metrics, which exhibit a generic unimodal distribution, the entropy of initial thought tokens ($\mathbf{H}_\text{init}$) displays a distinct bimodal and heavy-tailed distribution. This indicates that $\mathbf{H}_\text{init}$ serves as a discriminative signal, effectively capturing the ``Aha Moments'' that distinguish complex reasoning steps from routine derivations.
    }
    \label{fig:distribution}
    \vspace{-10pt}
\end{figure*}

\section{A Preliminary Study}
\label{sec:motivation}
Previous works have suggested that the onset of a reasoning step--often marked by discourse cues such as ``Wait'', ``But'', or ``So''--represents a cognitive pivot that shapes the entire subsequent reasoning trajectory~\cite{wang2025beyond,yang2025speculative,zeng2025pruning,shi2025longcodezip}. Building on this intuition, we hypothesize that the uncertainty associated with the \textit{initial token} of a reasoning step provides a more discriminative signal of step-level difficulty than statistics aggregated over the full step. In other words, the essential difficulty information is concentrated at the very beginning of each reasoning step. 
In this section, we analyze the uncertainty distribution for the reasoning steps and examine how different uncertainty metrics correlate with step-wise difficulty.

\subsection{Study Design}
\label{ssec:metrics}
To quantify uncertainty within a reasoning step, we analyze entropy-based metrics derived from the model’s conditional probability distribution $P_\theta$. It is well-established that the internal logits of LLMs serve as a reliable proxy for model confidence~\cite{malinin2020uncertainty, kuhn2023semantic}. 
Let \(s_k=\{t_1,\ldots,t_L\}\) denote the \(k\)-th reasoning step with \(L\) tokens, and \(\mathbf{c}_k\) be the preceding context of this step.

\paragraph{Step Entropy ($\mathbf{H}_\text{step}$).} $\mathbf{H}_\text{step}$ quantifies the average uncertainty across all tokens in a reasoning step:
\vspace{-20pt}
\begin{equation}
    \mathbf{H}_\text{step}(s_k) = \frac{1}{L} \sum_{i=1}^{L} \mathbf{H}(P_\theta(\cdot | t_{<i}, \mathbf{c}_k))
\end{equation}
where $\mathbf{H}(p) = - \sum_{v \in V} p(v) \log p(v)$ denotes the Shannon entropy of a distribution~\cite{shannon1948mathematical}. 
This metric treats all tokens within the step uniformly by averaging uncertainty across the entire sequence.

\paragraph{Initial Token Entropy ($\mathbf{H}_\text{init}$).} In contrast, \(\mathbf{H}_\text{init}\) captures the uncertainty of the \emph{initial token} for each step:
\begin{equation}
    \mathbf{H}_\text{init}(s_k) = \mathbf{H}(P_\theta(t_1 | \mathbf{c}_k))
\end{equation}
Unlike step-level averages, $\mathbf{H}_\text{init}$ isolates the uncertainty at the onset of the step, which we hypothesize to be an indicator of the step’s difficulty.


As additional baselines, we consider the following commonly used metrics in collaborative inference.

\paragraph{Step-wise Perplexity ($\mathbf{PPL}_\text{step}$).} $\mathbf{PPL}_\text{step}$ measures the exponentiated average negative log-likelihood over the entire step:
\begin{equation}
    \mathbf{PPL}_\text{step}(s_k) = \exp\left( -\frac{1}{L} \sum_{i=1}^{L} \log P_\theta(t_i | t_{<i}, \mathbf{c}_k) \right)
\end{equation}
Like $\mathbf{H}_\text{step}$, $\mathbf{PPL}_\text{step}$ aggregates uncertainty uniformly across all tokens.

\paragraph{LLM-as-a-Judge.} It assesses the validity of a reasoning step using the large model directly. Given a step $s_k$ and its context $\mathbf{c}_k$, the LLM assigns a scalar score $S \in \{0, 1, \dots, 9\}$.
We note that this metric requires an additional computationally expensive inference pass and is therefore unsuitable for efficiency-critical routing decisions.

We examine the reasoning traces using Qwen3-4B, Qwen3-32B~\cite{yang2025qwen3}, and DeepSeek-R1-Distill-Qwen-32B~\cite{guo2025deepseek} on the AIME and LiveCodeBench datasets~\cite{jain2024livecodebench}, collecting over 10 million tokens of reasoning steps. 


\begin{figure}[tbh]
    \centering
    \includegraphics[width=\columnwidth]{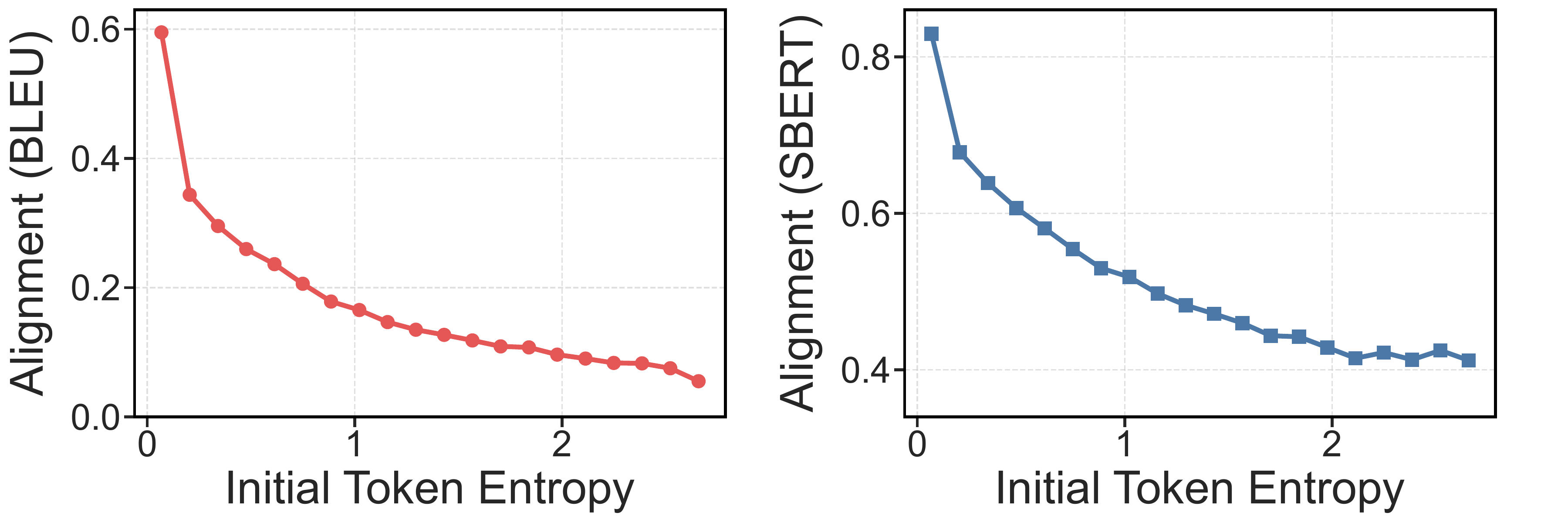}
    \caption{Alignment between the outputs generated by small and large models under various $\mathbf{H}_\text{init}$ intervals.} 
    \label{fig:similarity}
    \vspace{-10pt} 
\end{figure}

\begin{figure*}[tbh]
    \centering
    \includegraphics[width=\textwidth]{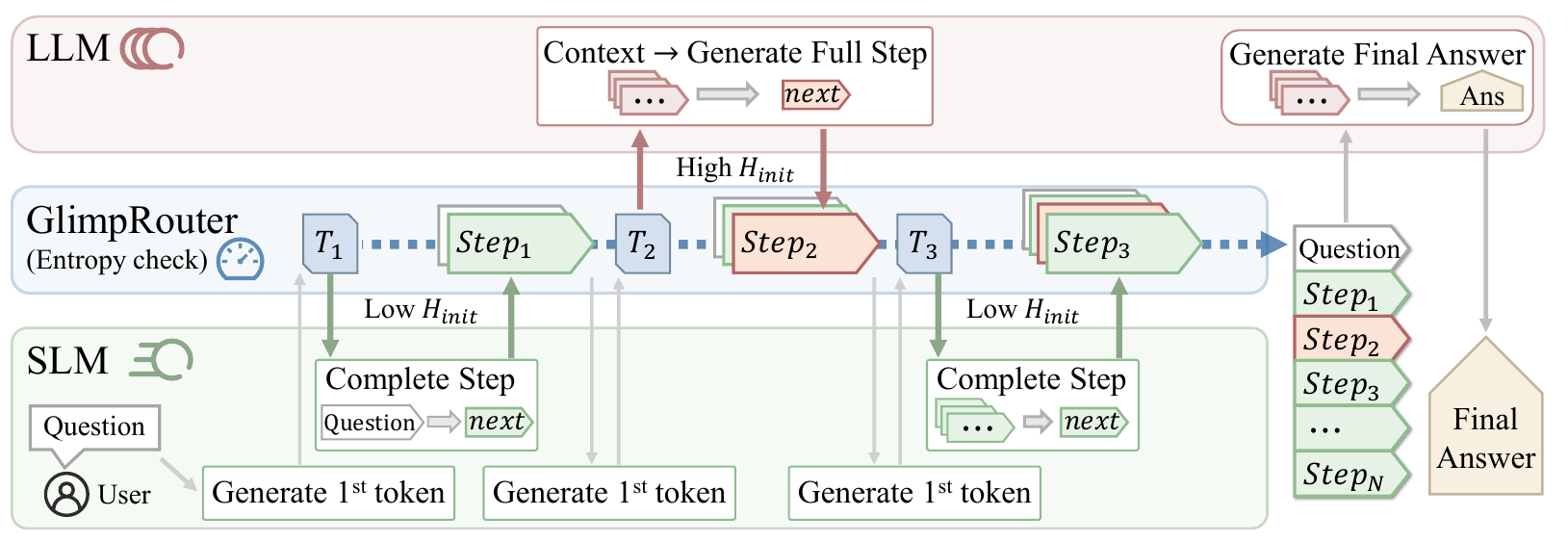}
    \caption{
The overall framework of \textbf{\approach}. The collaborative inference pipeline operates in a step-wise manner. At the onset of each reasoning step, the SLM first generates a ``glimpse'' (the initial token $T_k$). GlimpRouter then computes the entropy $\mathbf{H}_\text{init}$ of this token to gauge step difficulty. Based on the threshold, the system dynamically routes the generation: (1) Delegate (Low $\mathbf{H}_\text{init}$): Routine steps are fully generated by the efficient SLM. (2) Intervene (High $\mathbf{H}_\text{init}$): Complex steps signaling cognitive pivots are routed to the capable LLM for robust reasoning. Finally, the LLM generates the conclusive answer based on the aggregated collaborative chain.
}
    \label{fig:framework4}
    \vspace{-10pt}
\end{figure*}

\subsection{Results and Analysis}
\label{ssec:empirical_analysis}

\paragraph{Information Distribution within Reasoning Steps.} 
Figure~\ref{fig:distribution} illustrates the distributions reflected by the four metrics.
It is evident that both $\mathbf{PPL}_\text{step}$ and $\mathbf{H}_\text{step}$ exhibit narrow, unimodal distributions clustered around central values. We attribute this to the \emph{signal dilution} effect: a reasoning step typically contains a few critical decision tokens followed by a long sequence of deterministic syntactic tokens. Averaging over the entire step smooths out the local spikes of uncertainty, making these metrics insensitive to the actual difficulty.
Similarly, the LLM-as-a-Judge distribution is discrete and highly skewed towards high scores (saturation), lacking the granularity required for fine-grained routing thresholds. 
In contrast, $\mathbf{H}_\text{init}$ presents a distinct bimodal distribution with a heavy tail. The clear separation between the low-entropy peak (routine steps) and the high-entropy tail (complex reasoning) suggests that $\mathbf{H}_\text{init}$ naturally functions as a high-sensitivity discriminator for routing.

\paragraph{Correlations to Model Routing.}
Motivated by the observation that most information in a reasoning step resides at its onset, we further examine whether this signal can serve as an indicator for model routing, where a lightweight language model can reliably substitute a large model. To this end, we partition reasoning steps into intervals based on $\mathbf{H}_\text{init}$ predicted by the small model. For each interval, we evaluate the alignment between the outputs generated by the small and large models, conditioned on the same preceding context \(\mathbf{c}\). We quantify alignment using two widely used similarity metrics: BLEU-4~\cite{lin2004orange}, which captures lexical overlap, and SBERT~\cite{reimers-2019-sentence-bert}, which measures semantic similarity. 
As shown in Figure~\ref{fig:similarity}, we observe a strictly monotonic negative correlation between $\mathbf{H}_\text{init}$ and both similarity metrics. In the low-$\mathbf{H}_\text{init}$ regime, the small model’s outputs closely align with those of the large model, indicating that such steps can be handled competently by the lightweight model. In contrast, as $\mathbf{H}_\text{init}$ increases, the alignment degrades substantially, reflecting growing divergence in both surface form and semantic content. This strong correlation provides empirical evidence that $\mathbf{H}_\text{init}$ can be a reliable predictor of step-level difficulty, thereby suggesting a routing strategy in which a small model can substitute a large model in low-$\mathbf{H}_\text{init}$ steps without compromising results.
To provide concrete intuition for this correlation, we present qualitative case studies in Appendix~\ref{app:case_studies}.

\section{Methodology}
\label{sec:method}


\subsection{Problem Formulation}
\label{ssec:formulation}

We consider a collaborative inference framework involving a large, high-capacity reasoning model (LLM, $M_L$) and a small computationally efficient model (SLM, $M_S$). 
Large reasoning models typically output a reasoning process $\mathcal{T}$, often encapsulated within a \texttt{<think>} \dots \texttt{</think>} tag, followed by the final answer $A$. 
Our framework focuses on accelerating the reasoning process $\mathcal{T}$, while the final answer $A$ is always generated by $M_L$ to ensure correctness.
Formally, the reasoning process is decomposed into a sequence of steps $\mathcal{T} = \{s_1,\dots, s_K\}$ based on structural delimiters (e.g., double newlines~\cite{pan2025specreason, yang2025speculative}). Each step $s_k$ consists of a sequence of tokens $s_k = (t_1, \dots, t_L)$ and is generated conditioned on the preceding context $\mathbf{c}_k$, which includes the original question and all previous generated steps {\(s_1,\ldots,s_{k-1}\)}.
The objective of collaborative inference is to dynamically assign each reasoning step $s_k$ to $M_S$ or $M_L$, minimizing overall inference latency while preserving the quality of reasoning required for accurate solutions.



\subsection{Overview}
To navigate the efficiency-accuracy trade-off, we propose \textbf{\approach}, a training-free, step-aware collaboration strategy that routes reasoning steps based on the \textit{initial token entropy} ($\mathbf{H}_\text{init}$).
Figure~\ref{fig:framework4} illustrates the overall pipeline.
Given an input question, the system operates in a step-wise manner. At the onset of each step, instead of blindly generating the full content, \approach employs $M_S$ to ``glimpse'' the first token, yielding an entropy $\mathbf{H}_\text{init}$ that quantizes the difficulty of the upcoming step: 
If $\mathbf{H}_\text{init}$ falls below a threshold, the step is deemed routine, and the small model is responsible for continuing the generation.
Otherwise, the step signals a cognitive pivot, and the context is handed over to a large model for high-quality generation.
This ``Probe-then-Dispatch'' mechanism ensures that heavy computational resources are allocated solely for critical reasoning steps. Each step is introduced in the following sections.
The detailed procedure is outlined in Algorithm~\ref{alg:ite_router} in Appendix~\ref{app:alg}.

\subsection{Glimpse: Initial Token Probing} 
At the beginning of the reasoning step $k$, given the preceding context $\mathbf{c}_k$, the small model $M_S$ is invoked to predict the probability distribution of only the first token $t_{k,1}$. We compute the entropy $\mathbf{H}_\text{init}(s_k)$ of this distribution as a proxy for the cognitive uncertainty of the current reasoning step.
This probing operation incurs a marginal computational cost equivalent to decoding a single token, unlike methods that necessitate generating an entire reasoning step (typically $L \gg 1$ tokens) before verification. Even in scenarios where the step is subsequently identified as difficult and routed to $M_L$ (thereby discarding the probe), this 1-token overhead is negligible compared to the substantial sunk costs associated with discarding fully generated invalid steps. 

\subsection{Dynamic Model Routing} 
Given the entropy of the initial token in the current step, \approach dispatches the generation task to the appropriate model based on a threshold $\tau$:

\noindent \textbf{Delegate ($\mathbf{H}_\text{init}(s_k) \le \tau$):} Low entropy suggests that the small model is confident in the logical progression, and the step is likely routine. Consequently, $M_S$ continues to autoregressively generate the remainder of step $s_k$ until the delimiter is reached. This decision reduces the total cost by maximizing the utilization of the small model.

\noindent \textbf{Intervene ($\mathbf{H}_\text{init}(s_k) > \tau$):} High entropy indicates logical ambiguity or high cognitive load. In this scenario, $M_L$ is selected to generate the step $s_k$. While this incurs a higher computational cost, it leverages the superior reasoning and inherent self-correction capabilities of LRMs~\cite{guo2025deepseek}. Specifically, $M_L$ can rectify potential logical drifts accumulated in the historical context $\mathbf{c}_k$, thereby satisfying the quality constraint. We provide a qualitative analysis of this implicit self-correction behavior in Appendix~\ref{app:case_studies}. 

\subsection{Efficient Model Switching}
A critical requirement for a step-level collaboration system, where every step involves context transitions, is minimizing the model switching overhead. To minimize the system overhead, we leverage prefix caching mechanisms supported by inference engines~\cite{kwon2023efficient, zheng2024sglang}.  
When routing a request between models, the context $\mathbf{c}_k$, which comprises the question and historical steps, is largely resident in the KV cache from previous interactions. Thus, the context processing is reduced to a highly parallelizable prefill phase rather than a serial re-computation. The resulting switching latency is comparable to decoding a few tokens, ensuring that the computational savings from $M_S$ are not negated by routing overheads.

\subsection{Hierarchical Acceleration}
\label{ssec:hierarchical}


A distinct advantage of \approach lies in its step-level granularity. This coarse-grained design is inherently orthogonal to token-level optimizations, allowing our framework to be seamlessly integrated with various low-level acceleration techniques to achieve compound speedups. To maximize system throughput, we implement a hierarchical acceleration strategy in Section~\ref{ssec:specdecode}.
At the inter-step level, \approach acts as a global planner, assigning routine logical steps to $M_S$ to bypass the expensive $M_L$ entirely.
At the intra-step level, when $M_L$ is invoked, we further accelerate its generation using Speculative Decoding~\cite{leviathan2023fast}. Specifically, we employ a ``Draft-then-Verify'' pipeline where the small draft model $M_S$ proposes token sequences that are verified in parallel by $M_L$.

\begin{table*}[t]
\centering
\setlength{\tabcolsep}{3.5pt} 
\begin{tabular}{llcccccccccc}
\toprule
\multirow{2}{*}{\textbf{LLM}} & \multirow{2}{*}{\textbf{Method}} & \multicolumn{2}{c}{\textbf{AIME24}} & \multicolumn{2}{c}{\textbf{AIME25}} & \multicolumn{2}{c}{\textbf{GPQA}} & \multicolumn{2}{c}{\textbf{LCBv5}} & \multicolumn{2}{c}{\textbf{LCBv6}} \\
\cmidrule(lr){3-4} \cmidrule(lr){5-6} \cmidrule(lr){7-8} \cmidrule(lr){9-10} \cmidrule(lr){11-12}
 & & Acc $\uparrow$ & Lat $\downarrow$ & Acc $\uparrow$ & Lat $\downarrow$ & Acc $\uparrow$ & Lat $\downarrow$ & Acc $\uparrow$ & Lat $\downarrow$ & Acc $\uparrow$ & Lat $\downarrow$ \\
\midrule
- & SLM only & 48.33 & 99  & 45.00 & 105 & 61.11 & 76  & 47.90 & 100 & 43.71 & 94 \\
\midrule
\multirow{5}{*}{\textbf{Qwen3-32B}} 
& LLM only & 60.00 & 220 & 48.33 & 231 & 61.87 & 194 & \textbf{52.69} & 249 & \textbf{47.43} & 241 \\
& Random & 56.67 & 134 & 47.50 & 136 & 61.74 & 128  & 51.20 & 147 & 44.00 & 146 \\
& RSD      & 59.17 & 167 & 47.50 & 173 & 62.50 & 165 & 51.05 & 209 & 46.29 & 208 \\
& SpecCoT    & 58.33 & 161 & 48.33 & 170 & 61.62 & 163 & 51.05 & 195 & 45.71 & 192 \\
& SpecReason & 60.00 & 160 & 49.17 & 162 & 62.63 & 181 & 51.50 & 213 & 46.29 & 214 \\
& \approach & \textbf{60.83} & 145 & \textbf{51.67} & 147 & \textbf{63.01} & 142 & \textbf{52.69} & 162 & 47.14 & 165 \\

\midrule
\multirow{5}{*}{\textbf{DeepSeek-32B}} 
& LLM only & 57.50 & 197 & 46.67 & 220 & 61.62 & 176 & 52.40 & 219 & 46.86 & 214 \\
& Random & 56.67 & 137 & 48.33 & 147 & 62.50 & 114 & 52.25 & 142 & 47.00 & 131 \\
& RSD & 58.33 & 167 & 45.83 & 171 & 63.64 & 146 & 51.95 & 180 & 45.71 & 178 \\
& SpecCoT    & 60.00 & 159 & 50.00 & 175 & 62.75 & 135 & 53.29 & 168 & 47.14 & 165 \\
& SpecReason & 57.50 & 158 & 49.17 & 169 & 63.76 & 213 & 53.59 & 185 & 47.57 & 189 \\
& \approach & \textbf{60.83} & 143 & \textbf{51.67} & 163 & \textbf{64.02} & 129 & \textbf{54.64} & 160 & \textbf{48.29} & 160 \\

\bottomrule
\end{tabular}
\caption{
Performance comparison with Qwen3-32B and DeepSeek-R1-Distill-Qwen-32B as LLM, and SLM is fixed as Qwen3-4B. Acc and Lat represent Accuracy (Pass@1, \%) and Average Latency (s), respectively. The best performance within each group is highlighted in bold.
}
\label{tab:main_results}
\vspace{-10pt}
\end{table*}

\section{Experiments}
\label{sec:experiments}

\subsection{Experimental Setup}
\label{ssec:setup}

\paragraph{Models and Configurations.}
We conduct comprehensive experiments using the Qwen3~\cite{yang2025qwen3} and DeepSeek-R1~\cite{guo2025deepseek} families.
We utilize Qwen3-4B as the small model (SLM) and DeepSeek-R1-Distill-Qwen-32B as the large model (LLM). This setup serves as the default configuration for our results.
To verify the scalability of \approach, we extend our analysis to Qwen3-32B as the LLM in Section~\ref{sec:main_results} and DeepSeek-R1-Distill-Qwen-1.5B as the SLM in Appendix~\ref{app:scalability}, covering a spectrum of model sizes and architectural pairings.

\paragraph{Benchmarks.}
We evaluate our method on a diverse set of complex reasoning tasks.
For mathematical reasoning, we employ AIME24 and AIME25, which represent the frontier of mathematical problem-solving. 
For general reasoning, we use GPQA-Diamond~\cite{rein2024gpqa}, an expert-written, challenging multiple-choice dataset in biology, physics, and chemistry.
For code generation, we utilize LiveCodeBench (v5 and v6)~\cite{jain2024livecodebench}, which evaluates the model's ability to solve competitive programming problems.

\paragraph{Baselines.}
We compare \approach against standalone models and state-of-the-art collaborative inference baselines.
\textbf{Standalone SLM / LLM}: The individual performance of the small and large models, serving as the efficiency and performance boundaries, respectively.
\textbf{Random}: A method using a random score from 0 to 9 to select the model for inference.
\textbf{RSD}~\cite{liaoreward}: A reward-guided method that employs a trained Process Reward Model (PRM) to evaluate the quality of the step and determine whether to invoke the large model.
\textbf{SpecCoT}~\cite{shi2025speccot}: A selection-based collaboration method where the small model generates multiple candidate reasoning steps in parallel. The large model acts as a discriminator to select the optimal step.
\textbf{SpecReason}~\cite{pan2025specreason}: A verification-based method where the small model generates a step, which is then verified by the large model acting as a judge. Upon rejection, the large model falls back to generation.

\paragraph{Metrics.}
We assess the efficiency-performance trade-off using two key metrics.
\textbf{Pass@1 (Acc):} The percentage of problems solved correctly on the first attempt.
\textbf{Latency (Lat):} The end-to-end wall-clock time (in seconds) per question, serving as the primary indicator of inference efficiency.


\begin{figure*}[t]
    \centering
    \includegraphics[width=\textwidth]{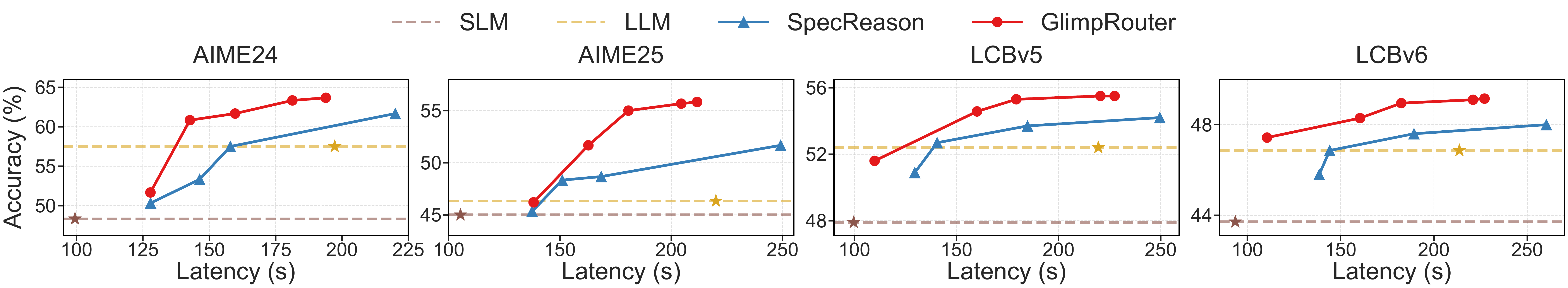}
    \caption{
    Sensitivity analysis of collaboration threshold on multiple benchmarks. \approach (red circles) establishes a superior Pareto frontier compared to the SpecReason (blue triangles). 
    Dashed lines represent the performance of standalone models.}
    \label{fig:tradeoff}
    \vspace{-10pt}
\end{figure*}

\paragraph{Implementation Details.}
All experiments are conducted on NVIDIA A100-80GB GPUs using the vLLM inference engine~\cite{kwon2023efficient}. We employ sampling with a temperature of $0.6$ and a top-p of $0.95$. The maximum budget for reasoning tokens is set to 8192.
For the hierarchical acceleration strategy, we repurpose the SLM to serve as the token-level drafter for the LLM with a draft length of $n=3$.
To ensure statistical stability, all reported results are averaged over 4 runs.
Regarding the entropy threshold $\tau$, since it modulates the trade-off between latency and accuracy, we report the configuration that achieves the optimal balance, maximizing accuracy while securing significant latency reduction, in our main results. Empirically, this corresponds to an \textit{intervention rate} (percentage of steps generated by LLM) of approximately 20\%--30\%. A detailed sensitivity analysis of $\tau$ is provided in Section~\ref{ssec:sensitive}.

\subsection{Main Results}
\label{sec:main_results}

Table~\ref{tab:main_results} presents the comparative performance of \approach against baselines on five complex reasoning benchmarks. 
\approach consistently achieves the optimal trade-off between inference latency and reasoning accuracy across different model configurations. In the primary setting using Qwen3-4B as the SLM, our method consistently outperforms the standalone LLM (DeepSeek-R1-Distill-Qwen-32B) in accuracy while delivering substantial speedups.
Specifically, \approach reduces end-to-end latency by \textbf{25.2\%--27.4\%} across all datasets compared to the LLM-only baseline. 
For instance, on AIME25, we observe a substantial \textbf{10.7\%} relative improvement in accuracy combined with a \textbf{25.9\%} reduction in latency. 

Compared to existing step-level collaboration methods, \approach demonstrates a distinct advantage in efficiency.
While baselines like RSD, SpecCoT, and SpecReason achieve better accuracy than the SLM-only baseline, they still incur significant latency overheads. 
For example, SpecReason's latency on GPQA (213s) exceeds even that of the standalone LLM, DeepSeek-R1-Distill-Qwen-32B (176s), negating the benefits of collaboration.
This inefficiency stems from their routing mechanisms: all methods require generating full reasoning steps (and often multiple candidates) or performing post-hoc verification before making a decision.
In contrast, \approach determines the routing path based solely on the \textit{initial token entropy}. This ``Probe-then-Dispatch'' mechanism minimizes the computational overhead of the decision process. 

A counterintuitive yet compelling result is that our collaborative framework outperforms the accuracy of the standalone large model (e.g., 51.67\% vs. 46.67\% on AIME25). 
We attribute this to the self-correction capacity of LRMs~\cite{guo2025deepseek, zeng2025thinking}.
The intervention mechanism in our framework offers more than just text completion. The high entropy at the onset of a step often serves as a manifestation of latent inconsistencies accumulated in the preceding trajectory. When \approach detects this uncertainty and invokes the LLM, the large model does not merely continue the sequence; effective intervention allows it to implicitly re-evaluate the context and rectify prior logical drifts, thereby realigning the reasoning path.
We provide concrete case studies illustrating this correction behavior in Appendix~\ref{app:case_studies}. 

\subsection{Ablation and Analysis}
\label{sec:ablation_study}

\paragraph{Sensitivity Analysis of Collaboration Threshold.}
\label{ssec:sensitive}

The trade-off between inference efficiency and reasoning capability is closely associated with the proportion of steps routed to the large model, referred to as the \textit{intervention rate}. To investigate how this rate impacts the performance, we vary the entropy threshold $\tau$ for \approach and the verification threshold for SpecReason. By varying these thresholds, we modulate the proportion of steps processed by the large model. Figure~\ref{fig:tradeoff} illustrates the accuracy-latency trade-off on multiple benchmarks. 
The curve of \approach (red) lies strictly above and to the left of the SpecReason baseline (blue), establishing a superior Pareto frontier. 
Detailed numerical results are provided in Appendix~\ref{app:sensitivity}.


\paragraph{Impact of Metric Choice.}
\label{ssec:metric_ablation}

To empirically validate the effectiveness of our $\mathbf{H}_\text{init}$ metric, we conducted an ablation study comparing it against two widely used uncertainty metrics: Step-wise Entropy ($\mathbf{H}_\text{step}$) and Step-wise Perplexity ($\mathbf{PPL}_\text{step}$).
Table~\ref{tab:metric_ablation} presents the comparative results on the AIME25 benchmark, while the extended results are detailed in Appendix~\ref{app:metric_ablation}.
\approach outperforms both step-wise variants by a significant margin, achieving a relative accuracy gain of \textbf{10.7\%} over $\mathbf{H}_\text{step}$ and \textbf{8.8\%} over $\mathbf{PPL}_\text{step}$.
This result corroborates our \textit{Signal Dilution} hypothesis (Section~\ref{ssec:empirical_analysis}). 
Beyond accuracy, $\mathbf{H}_\text{init}$ offers a fundamental advantage in latency (163s vs. 178s / 181s).
Implementing routing via step-wise metrics necessitates a ``Generate-then-Measure'' paradigm, where the initial generation becomes a sunk cost if the step is subsequently routed to $M_L$.
Conversely, $\mathbf{H}_\text{init}$ operates on a ``Probe-then-Dispatch'' basis, eliminating the overhead of generating invalid draft steps.

\begin{table}[t]
\centering
\setlength{\tabcolsep}{10pt}
\begin{tabular}{lcc}
    \toprule
    \textbf{Variants} & \textbf{Acc $\uparrow$} & \textbf{Lat $\downarrow$} \\
    \midrule
    \textbf{\approach} & \textbf{51.67} & \textbf{163} \\
    \quad w/ step-wise entropy & 46.67 & 178 \\
    \quad w/ step-wise perplexity & 47.50 & 181 \\
    \bottomrule
\end{tabular}
\caption{Ablation study of different metrics on AIME25. We report accuracy (Acc) and average generation latency (Lat) measured in seconds.}
\label{tab:metric_ablation}
\vspace{-10pt}
\end{table}


\paragraph{Orthogonal Speedup with Speculative Decoding.}
\label{ssec:specdecode}

To demonstrate the versatility of our framework, we evaluate the compatibility of \approach with \textbf{Speculative Decoding}, a token-level acceleration technique. Table~\ref{tab:orthogonality} presents the performance metrics when integrating Speculative Decoding into three different inference paradigms: standalone LLM, SpecReason, and \approach.
As shown in Table~\ref{tab:orthogonality}, integrating Speculative Decoding consistently reduces latency across all methods while maintaining comparable accuracy.
Crucially, the combination of \textbf{\approach + Speculative Decoding} achieves the lowest end-to-end latency among all configurations. 
(1) \textbf{Global Synergy:} \approach optimizes the coarse-grained logical flow by routing easy steps to the small model, reducing the total number of calls to the expensive large model.
(2) \textbf{Local Synergy:} For the difficult steps that are processed by LLM, Speculative Decoding optimizes the fine-grained token generation, mitigating the high per-token cost of the large model.
By attacking the efficiency bottleneck from both step-level routing and token-level execution, our framework achieves a compound speedup that surpasses either technique applied in isolation.
Additional results on other benchmarks are available in Appendix~\ref{app:orthogonality}.


\begin{table}[tbh]
\centering
\setlength{\tabcolsep}{0.5pt} 
\begin{tabular}{lcccccc}
\toprule
\multirow{2}{*}{\textbf{Method}} & \multicolumn{2}{c}{\textbf{AIME25}} & \multicolumn{2}{c}{\textbf{LCBv6}} \\
\cmidrule(lr){2-3} \cmidrule(lr){4-5}
& Acc $\uparrow$ & Lat $\downarrow$ & Acc $\uparrow$ & Lat $\downarrow$ \\
\midrule
LLM only         & 46.67 & 220 & 46.86 & 214 \\
\rowcolor{gray!20}
\quad + Speculative Decoding        & 45.83 & 149 & 46.29 & 166 \\
SpecReason          & 49.17 & 169 & 47.57 & 189 \\
\rowcolor{gray!20}
\quad + Speculative Decoding        & 49.17 & 140 & 47.14 & 154 \\
\approach          & 51.67 & 163 & 48.29 & 160 \\
\rowcolor{gray!20}
\quad + Speculative Decoding        & 51.67 & 130 & 48.00 & 137 \\
\bottomrule
\end{tabular}
\caption{Evaluation of orthogonality with Speculative Decoding on AIME25 and LCBv6. We report accuracy (Acc) and average generation latency (Lat) measured in seconds.} 
\label{tab:orthogonality}
\vspace{-10pt}
\end{table}

\section{Related Work}
\label{sec:related_work}

\paragraph{Collaborative Inference.}
Collaborative inference has emerged as a promising paradigm for optimizing the efficiency-performance trade-off by leveraging models of varying capacities. Existing approaches generally operate across three interaction granularities: query-level routing~\cite{chen2023frugalgpt, ding2024hybrid, ong2024routellm}, step-level collaboration~\cite{shi2025speccot, pan2025specreason, liaoreward}, and token-level speculation~\cite{leviathan2023fast, chen2023accelerating, cai2024medusa, li2024eagle}.
Our \approach distinguishes itself by proposing a training-free, step-level ``Probe-then-Dispatch'' mechanism based on initial token entropy, which effectively eliminates the sunk costs associated with post-hoc verification.

\paragraph{Efficient Reasoning.}
LRMs such as DeepSeek-R1~\cite{guo2025deepseek} and OpenAI o1/o3~\cite{jaech2024openai,openai2025o3} demonstrate that scaling test-time compute via Chain-of-Thought~\cite{wei2022chain} reasoning significantly enhances reasoning capabilities, albeit at the cost of prohibitive latency.
To mitigate this bottleneck, recent research explores dynamic offloading computation to smaller models~\cite{chen2025survey, wang2025mixllm,qu2025survey}.
At the query and token levels, methods utilize trained difficulty predictors~\cite{damani2024learning} or trained routers~\cite{fu2025r2r}.  
Most pertinent to logical deduction is step-level collaboration. Existing methods range from training-based reward guidance (RSD)~\cite{liaoreward} to training-free paradigms relying on multi-path selection (SpecCoT)~\cite{shi2025speccot} or post-hoc verification (SpecReason)~\cite{pan2025specreason}.
Unlike these methods constrained by redundant generation or heavy verification overheads, \approach achieves efficient orchestration via initial token probing.

\section{Conclusion}
\label{sec:conclusion}

In this work, we addressed the critical latency bottleneck of LRMs by proposing \approach. This training-free, step-wise collaborative inference framework orchestrates models of varying capacities.
We introduce the ``Probe-then-Dispatch'' mechanism based on initial token entropy. This approach allows the system to anticipate the step difficulty with negligible overhead.
Extensive experiments on multiple benchmarks demonstrate that \approach establishes a superior Pareto frontier. 
We hope this work enables efficient reasoning and stimulates research on dynamic computation allocation.

\section*{Limitations}

Despite the efficiency gains demonstrated by \approach, we acknowledge several limitations.
First, our routing mechanism relies on a static entropy threshold. While initial token entropy provides a more continuous and granular signal compared to discrete metrics like LLM-as-a-Judge, a fixed global threshold may not adapt optimally to the varying difficulty distributions across diverse domains or specific query types. Future work could explore adaptive or instance-aware thresholding mechanisms to further refine the efficiency.
Second, the step-level decomposition in our framework relies on explicit structural delimiters (specifically, double newline characters). Although this formatting pattern is prevalent in Large Reasoning Models, such as the DeepSeek-R1 and Qwen families, this heuristic dependence may limit the framework's direct applicability to models that generate unstructured Chain-of-Thought sequences. Exploring semantic-based segmentation strategies remains a valuable direction for future research.

\section*{Acknowledgments}

This research is funded by the National Key Research and Development Program of China (Grant No. 2023YFB4503802) and the Natural Science Foundation of Shanghai (Grant No. 25ZR1401175).



\bibliography{custom}


\appendix

\vspace{10pt}
\section{Pseudocode of \approach}
\label{app:alg}

The pseudocode of \approach is provided in Algorithm~\ref{alg:ite_router}.
Given an input question, the system operates in a step-wise manner. At the onset of each step, instead of blindly generating the full content, \approach employs $M_S$ to ``glimpse'' the first token, yielding an entropy $\mathbf{H}_\text{init}$ that quantizes the difficulty of the upcoming step: 
If $\mathbf{H}_\text{init}$ falls below a threshold, the step is deemed routine, and the small model is responsible for continuing the generation.
Otherwise, the step signals a cognitive pivot, and the context is handed over to a large model for high-quality generation.


        
        
        


\begin{algorithm}[tbh]
\small
\caption{Step-wise Collaborative Inference via \approach}
\label{alg:ite_router}
\begin{algorithmic}[1]
\State \textbf{Inputs:} question $q$, large model $M_L$, small model $M_S$, entropy threshold $\tau$
\State \textbf{Output:} a composite reasoning chain $\hat{\mathcal{T}}$

\State $\hat{\mathcal{T}} \leftarrow \emptyset$ // Initialize the reasoning chain 
\State $k \leftarrow 1$

\While{not finished}
    \State  $\mathbf{c}_k$=\textsc{ConCat}($q$, $\hat{\mathcal{T}}$) // Construct the current context by concatenating question and historical steps 
    \State $P_{\text{init}}$ = $M_S(\mathbf{c}_k)$ // Compute the first-token distribution using the small model 
    \State Calculate the initial token entropy $\mathbf{H}_\text{init}(s_k)$ 
    \If{$\mathbf{H}_\text{init}(s_k) > \tau$}
        \State $\hat{s}_k \leftarrow M_L(\mathbf{c}_k)$ // Generate $\hat{s}_k$ using the large model
    \Else
        \State $\hat{s}_k \leftarrow M_S(\mathbf{c}_k)$ // Generate $\hat{s}_k$ using the small model
    \EndIf
    
    \State $\hat{\mathcal{T}}$ = \textsc{ConCat}($\hat{\mathcal{T}}, \hat{s}_k$) // Append the generated step to the reasoning chain
    \State $k \leftarrow k + 1$
\EndWhile

\State \Return $\hat{\mathcal{T}}$
\end{algorithmic}
\end{algorithm}

\begin{table*}[t]
\centering
\setlength{\tabcolsep}{4.5pt} 
\begin{tabular}{lcccccccccc}
\toprule
\multirow{2}{*}{\textbf{Method}} & \multicolumn{2}{c}{\textbf{AIME24}} & \multicolumn{2}{c}{\textbf{AIME25}} & \multicolumn{2}{c}{\textbf{GPQA}} & \multicolumn{2}{c}{\textbf{LCBv5}} & \multicolumn{2}{c}{\textbf{LCBv6}} \\
\cmidrule(lr){2-3} \cmidrule(lr){4-5} \cmidrule(lr){6-7} \cmidrule(lr){8-9} \cmidrule(lr){10-11}
 & Acc $\uparrow$ & Lat $\downarrow$ & Acc $\uparrow$ & Lat $\downarrow$ & Acc $\uparrow$ & Lat $\downarrow$ & Acc $\uparrow$ & Lat $\downarrow$ & Acc $\uparrow$ & Lat $\downarrow$ \\
\midrule
SLM only & 25.83 & 60  & 26.67 & 57  & 15.91 & 37  & 15.12 & 66  & 20.29 & 62 \\
LLM only & 57.50 & 197 & 46.67 & 220 & 61.62 & 176 & 52.40 & 219 & 46.86 & 214 \\
\midrule
Random & 43.33 & 127 & 28.33 & 135 & 46.46 & 88 & 37.13 & 142 & 33.71 & 138 \\
RSD & 46.67 & 162 & 35.83 & 173 & 49.90 & 123 & 41.32 & 198 & 37.71 & 197 \\
SpecCoT    & 50.00 & 158 & 30.83 & 168 & 51.01 & 121 & 37.96 & 193 & 37.03 & 184 \\
SpecReason & 45.83 & 152 & 31.67 & 171 & 53.41 & 159 & 38.08 & 192 & 38.26 & 190 \\
\approach & \textbf{51.67} & 145 & \textbf{39.17} & 166 & \textbf{54.04} & 117 & \textbf{45.66} & 189 & \textbf{43.57} & 182 \\

\bottomrule
\end{tabular}
\caption{
Performance comparison with DeepSeek-R1-Distill-Qwen-1.5B as SLM and DeepSeek-R1-Distill-Qwen-32B as LLM. Acc and Lat represent Accuracy (Pass@1, \%) and Average Latency (s), respectively. The best performance is highlighted in bold.
}
\label{tab:ds_scalability}
\end{table*}

\section{Scalability across Architectural Pairings}
\label{app:scalability}

To verify the universality of \approach beyond specific small models (e.g., Qwen3-4B), we extend our evaluation to a homogeneous model family setting. Specifically, we employ DeepSeek-R1-Distill-Qwen-1.5B as the small model (SLM) and DeepSeek-R1-Distill-Qwen-32B as the large model (LLM). 

Table~\ref{tab:ds_scalability} presents the comparative results on the AIME, GPQA-Diamond, and LiveCodeBench benchmarks. 
Consistent with the observations in our main experiments (Section~\ref{sec:main_results}), \approach continues to exhibit a superior efficiency-performance trade-off compared to baselines.
Despite the change in the small model, \approach maintains its advantage over reactive baselines such as RSD, SpecReason, and SpecCoT.
While baselines incur significant latency overhead due to full-step generation and verification, \approach effectively leverages $\mathbf{H}_\text{init}$ to route simpler steps to DeepSeek-R1-Distill-Qwen-1.5B preemptively.
The success of \approach in this DeepSeek-R1-Distill-Qwen-1.5B + DeepSeek-R1-Distill-Qwen-32B setting underscores the robustness of the ``Probe-then-Dispatch'' mechanism. It suggests that the correlation between the initial token's entropy and step-level difficulty is not an artifact of specific datasets or model families, but rather an intrinsic property of large reasoning models.

\begin{table*}[t]
\centering
\setlength{\tabcolsep}{2pt} 
\begin{tabular}{ll ccc ccc ccc ccc}
\toprule
\multirow{2}{*}{\textbf{Method}} & \multirow{2}{*}{\textbf{Thr.}} & \multicolumn{3}{c}{\textbf{AIME24}} & \multicolumn{3}{c}{\textbf{AIME25}} & \multicolumn{3}{c}{\textbf{LCBv5}} & \multicolumn{3}{c}{\textbf{LCBv6}} \\
\cmidrule(lr){3-5} \cmidrule(lr){6-8} \cmidrule(lr){9-11} \cmidrule(lr){12-14}
 & & Acc $\uparrow$ & Lat $\downarrow$ & Rate & Acc $\uparrow$ & Lat $\downarrow$ & Rate & Acc $\uparrow$ & Lat $\downarrow$ & Rate & Acc $\uparrow$ & Lat $\downarrow$ & Rate \\
\midrule
\textbf{SLM only}  & - & 48.33 & 99 & 0 & 45.00 & 105 & 0 & 47.90 & 100 & 0 & 43.71 & 94 & 0 \\
\textbf{LLM only}  & - & 57.50 & 197 & 100 & 46.67 & 220 & 100 & 52.40 & 219 & 100 & 46.86 & 214 & 100 \\
\midrule
\multirow{4}{*}{\textbf{SpecReason}} 
 & $\sigma=6$ & 50.00 & 128 & 2 & 45.00 & 138 & 2 & 50.89 & 130 & 13 & 45.80 & 139 & 15 \\
 & $\sigma=7$ & 53.33 & 146 & 5 & 48.33 & 151 & 4 & 52.69 & 141 & 17 & 46.86 & 144 & 17 \\
 & $\sigma=8$ & 57.50 & 158 & 14 & 49.17 & 169 & 12 & 53.59 & 185 & 32 & 47.57 & 189 & 37 \\
 & $\sigma=9$ & 61.67 & 220 & 48 & 51.67 & 249 & 51 & 54.20 & 250 & 87 & 48.00 & 260 & 87 \\
\midrule
\multirow{5}{*}{\textbf{\approach}} 
 & $\tau=1.8$ & 51.67 & 128 & 2 & 45.83 & 138 & 2 & 51.60 & 110 & 2 & 47.43 & 111 & 2 \\
 & $\tau=0.9$ & 60.83 & 143 & 26 & 51.67 & 163 & 27 & 54.64 & 160 & 37 & 48.29 & 160 & 37 \\
 & $\tau=0.6$ & 61.67 & 160 & 44 & 55.00 & 181 & 45 & 55.30 & 179 & 60 & 48.95 & 183 & 65 \\
 & $\tau=0.1$ & 63.33 & 181 & 76 & 55.67 & 205 & 76 & 55.50 & 221 & 84 & 49.10 & 221 & 88 \\
 & $\tau=0.01$ & 63.67 & 194 & 85 & 55.83 & 212 & 83 & 55.50 & 227 & 89 & 49.15 & 227 & 91 \\
\bottomrule
\end{tabular}
\caption{
Detailed sensitivity analysis of routing thresholds across different benchmarks. 
\textbf{Thr.}: Routing threshold ($\tau$ for GlimpRouter, $\sigma$ for SpecReason). 
\textbf{Acc}: Pass@1 Accuracy. 
\textbf{Lat}: Average Latency.
\textbf{Rate}: Intervention Rate, the percentage of steps generated by the large model ($M_L$).
}
\label{tab:sensitivity_data}
\end{table*}

\section{Detailed Sensitivity Analysis of Collaboration Thresholds}
\label{app:sensitivity}

In this section, we present the comprehensive numerical results that underpin the sensitivity analysis discussed in Section~\ref{ssec:sensitive}. We examine the impact of varying the thresholds on the intervention rate, defined as the percentage of reasoning steps generated by the large model ($M_L$), and the resulting efficiency-performance trade-offs.
We evaluate two distinct routing mechanisms: 
We sweep the Initial Token Entropy ($\mathbf{H}_\text{init}$) threshold $\tau \in \{0.01, 0.1, 0.6, 0.9, 1.8\}$. A lower $\tau$ makes the system more cautious, triggering $M_L$ more frequently (Higher Intervention).
We vary the verification confidence threshold $\sigma \in \{6, 7, 8, 9\}$ of SpecReason. A stricter threshold leads to higher rejection rates, forcing $M_L$ to regenerate steps more frequently (Higher Intervention).

Table~\ref{tab:sensitivity_data} details the performance metrics on the AIME and LiveCodeBench benchmarks.
A critical observation is the scaling behavior of latency with respect to the intervention rate.
For \approach, latency increases linearly and modestly as the intervention rate rises. This is because the overhead of checking the first token is negligible; the latency cost is purely the difference between $M_L$ and $M_S$ generation speeds.
In stark contrast, SpecReason exhibits a super-linear latency spike as the intervention rate increases. Taking AIME25 as an example, at an intervention rate of approximately $\sim 50\%$, the latency of SpecReason is significantly higher than that of \approach. This confirms the ``Sunk Cost'' hypothesis: SpecReason must waste time generating a full draft by $M_S$ before $M_L$ can intervene, whereas \approach proactively dispatches the task, eliminating redundant computation.
Furthermore, comparing iso-accuracy configurations (e.g., \approach with \textbf{$\tau=0.9$} vs. SpecReason with \textbf{$\sigma=9$}, both achieving 51.67\% on AIME25), our method consistently achieves lower latency. This numerical evidence substantiates the visual trend observed in the Pareto frontier (Figure~\ref{fig:tradeoff}), demonstrating that $\mathbf{H}_\text{init}$ is more efficient than post-hoc verification.

\begin{table*}[t]
\centering
\setlength{\tabcolsep}{5pt} 
\begin{tabular}{lcccccccc}
\toprule
\multirow{2}{*}{\textbf{Variants}} & \multicolumn{2}{c}{\textbf{AIME24}} & \multicolumn{2}{c}{\textbf{AIME25}} & \multicolumn{2}{c}{\textbf{LCBv5}} & \multicolumn{2}{c}{\textbf{LCBv6}} \\
\cmidrule(lr){2-3} \cmidrule(lr){4-5} \cmidrule(lr){6-7} \cmidrule(lr){8-9}
 & Acc $\uparrow$ & Lat $\downarrow$ & Acc $\uparrow$ & Lat $\downarrow$ & Acc $\uparrow$ & Lat $\downarrow$ & Acc $\uparrow$ & Lat $\downarrow$ \\
\midrule
\textbf{\approach} & \textbf{60.83} & \textbf{143} & \textbf{51.67} & \textbf{163} & \textbf{54.64} & \textbf{160} & \textbf{48.29} & \textbf{160} \\
\quad w/ step-wise entropy & 58.33 & 163 & 46.67 & 178 & 53.44 & 185 & 47.29 & 194 \\
\quad w/ step-wise perplexity & 59.17 & 159 & 47.50 & 181 & 53.29 & 181 & 47.43 & 194 \\

\bottomrule
\end{tabular}
\caption{
Comprehensive ablation studies of metrics on four benchmarks. 
We report accuracy (Acc) and average generation latency (Lat) measured in seconds.
}
\label{tab:full_metric_ablation}
\end{table*}

\begin{table*}[t]
\centering
\setlength{\tabcolsep}{5pt} 
\begin{tabular}{lcccccccc}
\toprule
\multirow{2}{*}{\textbf{Method}} & \multicolumn{2}{c}{\textbf{AIME24}} & \multicolumn{2}{c}{\textbf{AIME25}} & \multicolumn{2}{c}{\textbf{LCBv5}} & \multicolumn{2}{c}{\textbf{LCBv6}} \\
\cmidrule(lr){2-3} \cmidrule(lr){4-5} \cmidrule(lr){6-7} \cmidrule(lr){8-9}
 & Acc $\uparrow$ & Lat $\downarrow$ & Acc $\uparrow$ & Lat $\downarrow$ & Acc $\uparrow$ & Lat $\downarrow$ & Acc $\uparrow$ & Lat $\downarrow$ \\
\midrule
LLM only & 57.50 & 197 & 46.67 & 220 & 52.40 & 219 & 46.86 & 214 \\
\rowcolor{gray!20}
\quad + Speculative Decoding & 56.67 & 139 & 45.83 & 149 & 51.80 & 165 & 46.29 & 166 \\
\midrule
SpecReason & 57.50 & 158 & 49.17 & 169 & 53.59 & 185 & 47.57 & 189 \\
\rowcolor{gray!20}
\quad + Speculative Decoding & 57.50 & 133 & 49.17 & 140 & 53.29 & 151 & 47.14 & 154 \\
\midrule
\approach & 60.83 & 143 & 51.67 & 163 & 54.64 & 160 & 48.29 & 160 \\
\rowcolor{gray!20}
\quad + Speculative Decoding & 60.00 & 116 & 51.67 & 130 & 54.34 & 134 & 48.00 & 137 \\
\bottomrule
\end{tabular}
\caption{
Comprehensive evaluation of orthogonal speedup with Speculative Decoding across four benchmarks. 
We report accuracy (Acc) and average generation latency (Lat) measured in seconds.
Rows highlighted in gray indicate the integration of token-level speculative decoding. 
} 
\label{tab:full_orthogonality}
\end{table*}

\section{Extended Analysis of Metric Choice}
\label{app:metric_ablation}

In Section~\ref{ssec:metric_ablation}, we demonstrate the superiority of the initial token entropy ($\mathbf{H}_\text{init}$) on the AIME25 benchmark. We extend the ablation study to include AIME24, AIME25, LiveCodeBench v5, and LiveCodeBench v6.
Table~\ref{tab:full_metric_ablation} presents the comprehensive results comparing \approach against Step-wise Entropy ($\mathbf{H}_\text{step}$) and Step-wise Perplexity ($\mathbf{PPL}_\text{step}$).

Across all four benchmarks, \approach employing $\mathbf{H}_\text{init}$ consistently achieves the highest accuracy.
This universal superiority strongly supports the \textbf{Signal Dilution} hypothesis. Step-wise metrics, by averaging uncertainty over the entire sequence length $L$, allow high-confidence syntactic tokens to obscure the signals from critical reasoning nodes.
In contrast, $\mathbf{H}_\text{init}$ focuses exclusively on the cognitive pivot (the initial token), providing a sharper, undiluted signal that effectively distinguishes between routine steps and complex logical bifurcations.
Regarding efficiency, the results highlight a fundamental structural advantage of our method.
On every benchmark, the step-wise metrics incur higher latency than \approach, despite operating at similar intervention rates.
This is because step-wise metrics operate on a ``Generate-then-Measure'' paradigm: the small model must fully generate a step before its difficulty can be assessed. If the step is subsequently rejected and routed to the large model, the initial generation becomes a sunk cost.
\approach, leveraging the ``Probe-then-Dispatch'' mechanism, assesses difficulty before generation. This proactive routing eliminates invalid drafting, ensuring that computational resources are allocated with maximal efficiency.

\section{Extended Analysis of Orthogonality with Speculative Decoding}
\label{app:orthogonality}

In Section~\ref{ssec:specdecode}, we demonstrated the compatibility of \approach with Speculative Decoding on specific benchmarks. To verify that this compound speedup is a universal property of our framework, we extend the evaluation to the full suite of benchmarks: AIME24, AIME25, LiveCodeBench v5, and LiveCodeBench v6.
Table~\ref{tab:full_orthogonality} details the performance metrics for three inference paradigms (Standalone LLM, SpecReason, and \approach), both with and without Token-level Speculative Decoding.

As evidenced in Table~\ref{tab:full_orthogonality}, integrating Speculative Decoding yields consistent latency reductions across all methods and benchmarks.
For the standalone LLM, latency drops by approximately \textbf{22\%--32\%}.
For the collaborative methods (SpecReason and \approach), the token-level acceleration further compresses the execution time of the large model's generation phases.
Importantly, this speedup is achieved with negligible impact on reasoning accuracy.
Crucially, the combination of \textbf{\approach + Speculative Decoding} achieves the lowest end-to-end latency across all experimental configurations.
Even when the strong baseline, SpecReason, is accelerated by Speculative Decoding, it remains consistently slower than our enhanced framework (e.g., 140s vs. 130s on AIME25).
This gap highlights a fundamental architectural distinction:
Token-level speculation can only accelerate the generation process; it cannot recover the sunk costs of invalid steps discarded during reactive verification.
SpecReason still pays the penalty for generating and rejecting full steps.
In contrast, \approach leverages $\mathbf{H}_\text{init}$ to proactively route steps, minimizing the number of tokens that need to be generated in the first place. When combined with Speculative Decoding, which accelerates the tokens that must be generated by $M_L$, we achieve a multiplicative effect on efficiency.
This confirms that \approach serves as an ideal \textbf{Global Planner}, perfectly complementing \textbf{Local Executors} like Speculative Decoding to establish a new state-of-the-art Pareto frontier.

\section{Case Studies}
\label{app:case_studies}

\subsection{Case Study 1: Correlation between Initial Token Entropy (\texorpdfstring{$\mathbf{H}_\text{init}$}{H init}) and Reasoning Difficulty}

We provide a concrete example to show the correlation between $\mathbf{H}_\text{init}$ and reasoning difficulty. The case visualizes how \approach dynamically allocates computational resources based on the cognitive load of each step.
The case below illustrates a representative trajectory from the AIME benchmark. We observe a clear dichotomy in $\mathbf{H}_\text{init}$ values corresponding to the nature of the reasoning steps:
\begin{itemize}
    \item \textbf{Low $\mathbf{H}_\text{init}$ (Routine Execution):} Steps 1-2 and 5-12 exhibit extremely low $\mathbf{H}_\text{init}$ values (e.g., $0.0046$, $0.0008$). These steps involve factual restatements or mechanical algorithmic operations (e.g., recursive division by 2). The small model ($M_S$) demonstrates high confidence and generates these tokens accurately, validating our ``Delegate'' strategy.
    \item \textbf{High $\mathbf{H}_\text{init}$ (Cognitive Bifurcation):} A critical spike in entropy occurs at the onset of Step 3 ($\mathbf{H}_\text{init}$: $1.8985$). Here, the reasoning process transitions from problem setup to solution planning. The small model's draft (starting with ``Maybe'') reflects uncertainty in selecting the counting method. 
\end{itemize}

\begin{tcolorbox}[
    breakable,
    colback=white,       
    colframe=gray!60,    
    boxrule=0.8pt,       
    arc=2mm,
    fonttitle=\bfseries,
    coltitle=black,
    title={A Case of Correlation between Initial Token Entropy ($\mathbf{H}_\text{init}$) and Reasoning Difficulty}
]
\small

\textbf{Question:} \\
Alice chooses a set $A$ of positive integers. Bob lists all finite nonempty sets $B$ whose maximum is in $A$. Bob's list has $2024$ sets. Find the sum of the elements of $A$.

\vspace{0.2cm}
\hrule
\vspace{0.2cm}

\textbf{Reasoning Trace via \approach:}

\noindent \textbf{Step 1} \texttt{[SLM]} {\scriptsize ($\mathbf{H}_\text{init}$: \textcolor{teal}{0.2369})} \\
First, Alice chooses a set $A$, and Bob lists all nonempty sets $B$ whose maximum is in $A$; there are $2024$ such sets.

\noindent \textbf{Step 2} \texttt{[SLM]} {\scriptsize ($\mathbf{H}_\text{init}$: \textcolor{teal}{0.3485})} \\
If $A=\{2\}$, then $B$ can be $\{2\}$ or $\{1,2\}$, giving $2^{2-1}=2$ sets. Generally, the number of sets corresponds to a sum of powers of 2.

\vspace{0.1cm}
\noindent \colorbox{red!10}{\parbox{\dimexpr\linewidth-6pt}{
\textbf{Step 3 (Critical Pivot)} \\
\textbf{Content ($M_S$):} ``Maybe I can try to find...'' {\scriptsize ($\mathbf{H}_\text{init}$: \textcolor{red}{\textbf{1.8985}} $> \tau$)} \\
$\hookrightarrow$ \textbf{Action:} \textcolor{red}{\textbf{Intervene ($M_L$)}} \\
\textbf{Content ($M_L$):} The most efficient way is to divide $2024$ by $2$ repeatedly to find its binary representation directly.
}}
\vspace{0.1cm}

\noindent \textbf{Step 4} \texttt{[SLM]} {\scriptsize ($\mathbf{H}_\text{init}$: \textcolor{teal}{0.0046})} \\
Compute $2024/2=1012$, remainder $0$.

\noindent \textbf{Step 5} \texttt{[SLM]} {\scriptsize ($\mathbf{H}_\text{init}$: \textcolor{teal}{0.0008})} \\
Then $1012/2=506$, remainder $0$.

\noindent \textbf{Step 6} \texttt{[SLM]} {\scriptsize ($\mathbf{H}_\text{init}$: \textcolor{teal}{0.0102})} \\
Next $506/2=253$, remainder $0$.

\noindent \textbf{Steps 7-10} \texttt{[SLM]} {\scriptsize ($\mathbf{H}_\text{init}$ values: \textcolor{teal}{0.0055, 0.0032, 0.0041})} \\
\textit{[...Routine division steps: $253 \to 126 \to 63 \to 31$...]}

\noindent \textbf{Step 11} \texttt{[SLM]} {\scriptsize ($\mathbf{H}_\text{init}$: \textcolor{teal}{0.1205})} \\
So the binary representation is $11111101000_2$. This corresponds to powers $2^{10}, \dots, 2^3$. Since the term is $2^{a-1}$, the elements are $a \in \{11, 10, 9, 8, 7, 6, 4\}$.

\noindent \textbf{Step 12} \texttt{[SLM]} {\scriptsize ($\mathbf{H}_\text{init}$: \textcolor{teal}{0.0020})} \\
Finally, the sum is $11+10+9+8+7+6+4 = \boxed{55}$.

\end{tcolorbox}


\subsection{Case Study 2: Self-Correction Mechanism in LRMs}

The case below demonstrates the system's ability to detect and correct potential logical errors during the reasoning process. 
In combinatorics problems involving grid paths, an accurate definition of geometric properties is a prerequisite for calculation.
\begin{itemize}
    \item \textbf{Logical Instability:} In Step 3, the small model establishes the relationship between ``direction changes'' and ``path segments''. It produces content containing a factual error: equating four direction changes to \textit{four} segments (the correct number is $4+1=5$).
    \item \textbf{Correction via Intervention:} Upon triggering the high-entropy threshold in Step 4, \approach invokes the large model ($M_L$). Crucially, $M_L$ does not merely complete the SLM's erroneous sentence. Instead, it re-evaluates the context and generates a correct premise: ``that means it has five straight segments.'' 
    \item \textbf{Outcome:} By correcting the hallucinated step and injecting the correct geometric reasoning, $M_L$ steers the trajectory back to a valid path. Subsequently, $M_S$ successfully performs the routine combinatorial calculations (Steps 5-6) based on the corrected logic.
\end{itemize}


\begin{tcolorbox}[
    colback=white,
    colframe=gray!60,
    boxrule=0.8pt,
    arc=2mm,
    fonttitle=\bfseries,
    coltitle=black,
    title={A Case of the Self-correction Mechanism in LRMs}
]
\small

\textbf{Question:} \\
Consider the paths of length $16$ that follow the lines from the lower left corner to the upper right corner on an $8\times 8$ grid. Find the number of such paths that change direction exactly four times.

\vspace{0.2cm}
\hrule
\vspace{0.2cm}

\textbf{Reasoning Trace via \approach:}

\noindent \textbf{Step 1} \texttt{[SLM]} {\scriptsize ($\mathbf{H}_\text{init}$: \textcolor{teal}{0.2457})} \\
So, the problem is about paths on a grid where you start at (0,0) and end at (8,8), moving only right (R) or up (U), each step being a unit move.

\noindent \textbf{Step 2} \texttt{[SLM]} {\scriptsize ($\mathbf{H}_\text{init}$: \textcolor{teal}{0.3914})} \\
Let me recall that in such grid paths, a path that goes from (0,0) to (8,8) can be represented as a sequence of right (R) and up (U) moves.

\vspace{0.1cm}
\noindent \colorbox{red!10}{\parbox{\dimexpr\linewidth-6pt}{
\textbf{Step 3 \texttt{[SLM]} (Logical Divergence)} \\
Suppose a path starts... four direction changes mean the path has \textit{four segments}. \\
\textbf{Step 4 \texttt{[LLM]} (Logical Correction)} \\
Wait, if a path changes direction exactly four times, that means it has \textit{five straight segments}. Because each change of direction adds a new segment. For example, starting with R gives R-U-R-U-R.
}}
\vspace{0.1cm}

\noindent \textbf{Step 5} \texttt{[SLM]} {\scriptsize ($\mathbf{H}_\text{init}$: \textcolor{teal}{0.4949})} \\
Let me recall that the number of compositions of n into k positive integers is C(n-1, k-1). Because it's equivalent to placing k-1 dividers in the n-1 gaps between n items.

\noindent \textbf{Step 6} \texttt{[SLM]} {\scriptsize ($\mathbf{H}_\text{init}$: \textcolor{teal}{0.2771})} \\
Therefore, the total number is $2 \times [\binom{7}{2} \times \binom{7}{1}] = 2 \times 21 \times 7 = \boxed{294}$.

\end{tcolorbox}

\end{document}